%% file: journal.tex
\documentclass[journal]{journal}
\usepackage[bookmarks=false,hidelinks]{hyperref}

\usepackage{xcolor}
\usepackage{graphicx}
\usepackage[normalem]{ulem}
\usepackage{subcaption}
\usepackage{movie15}
\usepackage{wrapfig} 
\usepackage{booktabs} 
\usepackage{amsfonts}

\usepackage{algorithm} 
\usepackage{algpseudocode}

\newcount\ppnum
\newcommand\ppnumber[1]{%
        \ppnum=#1\relax
        \ifnum\ppnum<0
                $-$%
                \ppnum=-\ppnum
        \fi
        \let\pptemp\empty
        \loop\ifnum\ppnum>999
                \count255=\ppnum
                \divide\ppnum by1000
                \count255=\numexpr \count255 - 1000*\ppnum \relax
                \edef\pptemp{,\ifnum\count255<100 0\ifnum\count255<10 0\fi\fi
                             \the\count255 \pptemp}%
        \repeat
        \the\ppnum
        \pptemp
}

\newcommand{\fingering}{\textsc{piano-fingering}}
\newcommand{\pig}{\textsc{PIG}}

\def\pieces{90}
\def\notes{155107}

\newcount\a\a=\number\notes
\newcount\b\b=\number\pieces
\divide\a by\b
\def\averagenotes{\the\a}

\newcommand{\dataset}{\textsc{APFD}}

\ifCLASSINFOpdf
\else
\fi
\begin{document}
%
\title{At Your Fingertips: \\Extracting Piano Fingering Instructions from Videos} 
%
%
%

\author{Amit~Moryossef,
        Yanai~Elazar,
        and~Yoav~Goldberg}

%
%

\markboth{Journal of \LaTeX\ Class Files,~Vol.~6, No.~1, January~2007}%
{Shell \MakeLowercase{\textit{et al.}}: Bare Demo of IEEEtran.cls for Journals}
%



\maketitle
\thispagestyle{empty}

\begin{abstract} 
Piano fingering---knowing which finger to use to play each note in a musical piece, is a hard and important skill to master when learning to play the piano. While some sheet music is available with expert-annotated fingering information, most pieces lack this information, and people often resort to learning the fingering from demonstrations in online videos. We consider the AI task of automating the extraction of fingering information from videos. This is a non-trivial task as fingers are often occluded by other fingers, and it is often not clear from the video which of the keys were pressed, requiring the synchronization of hand position information and knowledge about the notes that were played. We show how to perform this task with high-accuracy using a combination of deep-learning modules, including a GAN-based approach for fine-tuning on out-of-domain data. We extract the fingering information with an f1 score of 97\%. We run the resulting system on \pieces~videos, resulting in high-quality piano fingering information of 150K notes, the largest available dataset of piano-fingering to date.
\end{abstract}


%
\IEEEpeerreviewmaketitle

\input{11_intro.tex}

\input{21_backgroung.tex}

\input{31_method.tex}


\input{41_results.tex}

\input{51_discussion.tex}

\ifCLASSOPTIONcaptionsoff
  \newpage
\fi



\bibliographystyle{IEEEtran}
%


\bibliography{piano}

\appendices

\include{999_appendix}
%








\end{document}

%% file: 11_intro.tex
\section{Introduction}
Learning to play the piano is a difficult task which takes years to master. 
One of the challenging aspects when learning a new piece is the fingering choice, in which the player has to choose what finger to use for each note. 
While beginner booklets contain many fingering suggestions, advanced pieces seldom do.

Automatic prediction of \fingering{} can be a useful tool for piano learners, to ease the learning process of new pieces. 
As manually labeling fingerings for different sheet music is an exhausting and expensive task\footnote{\cite{nakamura2019statistical} reported expert labeling time of 3-12 seconds per note.}, previous work \cite{parncutt1997ergonomic,hart:00,jacobs2001refinements,Kasimi2007ASA,nakamura2019statistical} used very few tagged pieces for evaluation, with minimal or no training data. 

\begin{figure}[t]
\centering
\begin{subfigure}{\linewidth}
    \includegraphics[width=\linewidth]{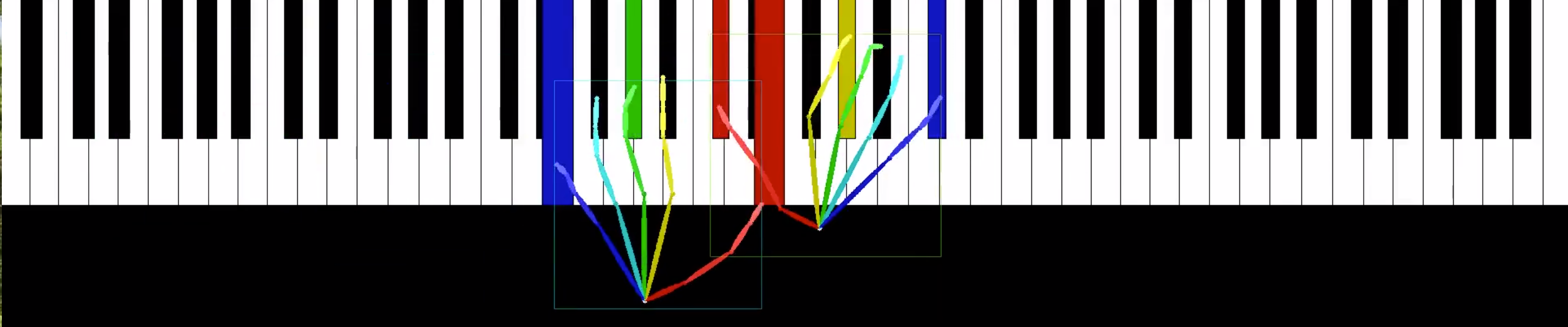}
    \caption{A recreation of the piano, hand pose estimation and predicted fingering, which are colored in corresponding colors to the fingers.}
    \label{fig:finger_detection:clean}
\end{subfigure}
\hfill
\begin{subfigure}{\linewidth}
    \includegraphics[width=\linewidth]{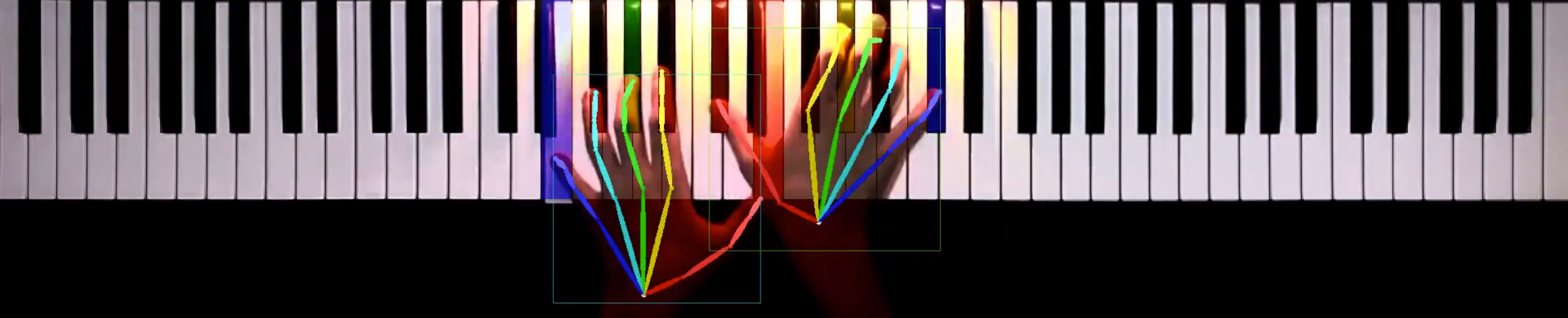}
    \caption{Overlay of what we perceive on the original frame.}
    \label{fig:finger_detection:overlay}
\end{subfigure}
\begin{subfigure}{0.48\linewidth}
    \begin{center}
    \includegraphics[height=1.5cm]{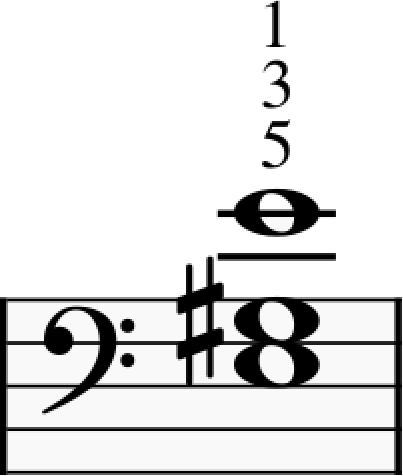}
    \caption{Left-hand sheet-music.}
    \end{center}
    \label{fig:finger_detection:f_clef}
\end{subfigure}
\begin{subfigure}{0.48\linewidth}
    \begin{center}
    \includegraphics[height=1.5cm]{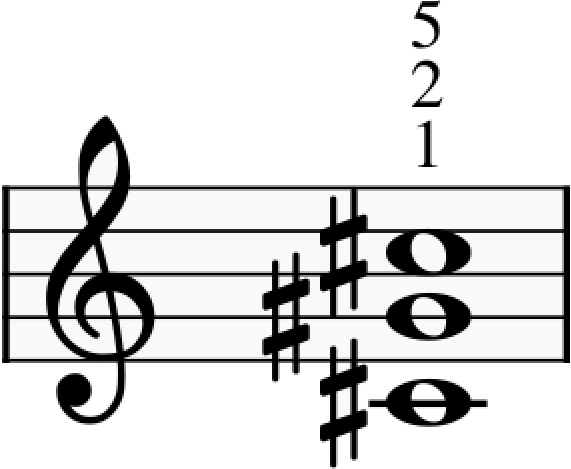}
    \caption{Right-hand sheet-music.}
    \end{center}
    \label{fig:finger_detection:g_clef}
\end{subfigure}
\caption{Illustration of the overall system output. We manage to automatically detect which fingers pressed each key of the piano.}
\label{fig:finger_detection}
\end{figure}

In this paper, we propose an automatic, low-cost method for detecting \fingering{} from piano playing performances captured on video, which allows training of modern, data-hungry neural networks. 
We introduce a novel pipeline that adapts and combines several deep learning methods such as Object-Detection \cite{viola2001rapid,redmon2016you,lin2017feature,lin2017focal}, Pose-Estimation \cite{wei2016convolutional,simon2017hand,DBLP:journals/corr/abs-1903-01013}, and GANs \cite{goodfellow2016deep,CycleGAN2017}, and results in an automatically labeled \fingering{} dataset. 
Our method serves two purposes: (1) an automatic ``transcript'' method that detects \fingering{} from video and MIDI files, when these are available, (Section \S\ref{sec:results:press-estimation}), and (2) running the method on dozens of videos, it yields a large dataset for training models genralizing to new pieces (Section \S\ref{sec:results:fingering-prediction}).

Given a video and a MIDI file, our system produces a probability distribution over the fingers for each played note.
Running this process on large corpora of piano performances played by different artists yields a total of \pieces{} automatically finger-tagged pieces (containing \ppnumber{\notes}~notes in total) resulting in the first public large scale \fingering{} dataset, which we name ``\textit{Automatic Piano Fingering Dataset} (\dataset)''. 
This dataset can grow over time, as more videos are uploaded to YouTube.
We provide empirical evidence that \dataset{} is valuable, by manually inspecting its labels quality and its usefulness for predictions by fine-tuning a model on a manually created dataset, which achieves state-of-the-art results on a \fingering{} task (\S\ref{sec:results:fingering-prediction}).

The process of extracting \fingering~from videos alone is difficult as it needs to detect keyboard presses, which are often subtle for the naked eye.
We therefore turn to MIDI files which were simultaneously recorded with videos to obtain this information. 
The extraction process works as follows:
We begin by creating a virtual keyboard, and map each key to match their location on the video (\S\ref{sec:dataset:keyboard}). 
Then, we identify the hands (\S\ref{sec:dataset:hand-detection}), and the fingers given the hands' bounding boxes (\S\ref{sec:dataset:pose}) for each frame of the video. 
Next, we align between the MIDI file and its corresponding video (\S\ref{sec:dataset:align}) and finally, for every played note, we assign a distribution over the fingers that played it (\S\ref{sec:dataset:press}). 

Even though methods for hand detection and pose estimation were extensively studied in the computer-vision literature \cite{kolsch2004robust,tang2015real,wei2016convolutional,simon2017hand,DBLP:journals/corr/abs-1903-01013}, we find that in practice, state-of-the-art models do not generalize, and perform poorly in our scenario. 
We address these weaknesses by fine-tuning an object detection model (\S\ref{sec:dataset:hand-detection}) on a new dataset that we manually annotated, and introduce a novel algorithm for adapting models on out-of-domain data.


We release this new hands dataset and the trained model in \cite{moryossef2021pianohands}, alongside the rest of the resources developed in this work, and a demonstration of the output from the entire process in \url{https://youtu.be/Gfs1UWQhr5Q}.

%% file: 21_backgroung.tex
\section{Background}\label{sec:background}

In this section, we present a literature review on the \fingering{} task, which was studied for many years. A shortcoming of most of these works is the lack of actual data for the different methods evaluation, which we aim to remedy by using our automatic method on videos.

\fingering{} was previously studied in multiple disciplines, such as music theory and computer animation \cite{parncutt1997ergonomic,hart:00,jacobs2001refinements,Kasimi2007ASA,zhu2013system,nakamura2019statistical}. Early works have mainly focused on search-based algorithms, with limited amounts of data and many unrealistic constraints. A recent work \cite{nakamura2019statistical} is the first work to produce a reasonable-sized corpus of piano fingering, with rigorous labeling by multiple professional players. Although the size of this dataset is enough for training modern statistical models, it is still considered relatively small, and more data can significantly boost performance.

The fingering prediction task is formalized as follows: Given a sequence of notes, associate each note with a finger from the set $\{1,2,3,4,5\} \times \{L,R\}$. This is subject to constraints such as the positions of each hand, anatomical plausibility of transitioning between two fingers, the hands' size, etc. Each fingering sequence has a cost of ``effort'' which is desired to be minimized.

One naive approach to finding the best optimal fingering sequence to play all the notes with is to exhaustively evaluate all possible transitions between one note to another which is not computationally feasible.
By defining a transition matrix corresponding to the probability or ``effort'' of transitioning from one note to another - one can calculate a cost function, which specifies the predicted sequence likelihood. Using a search algorithm on top of the transitions allows to find a globally optimal solution. But this solution is not practical, due to the exponential complexity, therefore heuristics and pruning are employed to reduce the space complexity.
The transition matrix can be manually defined by heuristics or personal estimation \cite{parncutt1997ergonomic,hart:00}, or instead of relying on a pre-defined set of rules, a Hidden Markov Model (HMM) can be used to learn the transitions \cite{yonebayashi:07,nakamura2014merged}. In practice, \cite{yonebayashi:07} leave the parameter learning to future work, and instead they manually fine-tune the transition matrix. 

On top of the transition matrix, practitioners suggested using dynamic programming algorithms to solve the search \cite{hart:00}. Another option to solve the huge search space is to use a search algorithm such as Tabu search \cite{glover1998tabu} or variable neighborhood search \cite{mladenovic1997variable}, to find a global plausible solution \cite{balliauw2015generating,balliauw2017variable}. 
These works are either limited by the defined transition rules, or by making different assumptions to facilitate the search space. Such assumptions come in the form of limiting the predictions to a single hand, limiting the evaluation pieces to contain no chords, rests or substantial lengths during which player can take their hand off the keyboard.
Furthermore, all of these works have small evaluation sets, which in practice makes it hard to compare different approaches, and does not allow to use more advanced models such as neural networks.

Previous works also constrained their models with manually defined rules that try to imitate human behaviour as these methods require lots of training data to achieve good performance, and fingering labeling is particularly expensive and hard to obtain. 
One way to automatically gather rich fingering data with full hand pose estimation is by using motion capture (MOCAP) gloves when playing the piano. \cite{zhu2013system} suggest a rule-based and data-based hybrid method, initially estimating fingering decisions using a Directed Acyclic Graph (DAG) based on rule-based comfort constraints which are smoothed using data recorded from limited playing sessions with motion capture gloves.
As MOCAP requires special equipment and may affect the comfort of the player, other work, \cite{takegawa2006design} tried to automatically detect piano fingering from video and MIDI files. The pianist's fingernails were laid with colorful markers, which were detected by a computer vision program. 
As some occlusions can occur, they used some rules to correct the detected fingering. In practice, they implemented the system with a camera capturing only 2 octaves (out of 8) and performed a very limited evaluation. The rules they used are simple (such as: restricting one finger per played note, two successive notes cannot be played with the same finger), but far from capturing real-world scenarios.

Previous methods for automatically collecting data \cite{takegawa2006design,zhu2013system} were costly, as apart of the equipment needed to play a piece, the data-collectors had to pay the participating pianists. In our work, we rely solely on videos from YouTube, meaning costs remain minimal with the ability to easily scale up to new videos.

Recently, \cite{nakamura2019statistical} released a relatively large dataset of manually labeled \fingering~by one to six annotators, consisting of 150 pieces, with partially annotated scores (324 notes per piece on average) with a total of 48,726 notes matched with 100,044 tags from multiple annotators, named \pig{}. This is the largest annotated \fingering{} corpus to date and a valuable resource for this field. The authors propose multiple methods for modeling the task of \fingering{}, including HMMs and neural networks, and report the best performance with an HMM-based model. 
In this work, we use \pig{} as a gold standard for comparison and show how a baseline model pre-trained on our dataset, can improve and achieve state-of-the-art results when fine-tuned on \pig{}.

%% file: 31_method.tex
\section{Our Approach: Extracting Fingering from Online Videos}
\label{sec:collection}
There is a genre of online videos in which people upload piano performances where both the piano and the hands are visible. On some channels, people not only include the video but also the MIDI file recorded while playing the piece.
We propose an algorithm for detecting the piano-keyboard, and the fingers playing the different notes, and estimating for each note the fingers that pressed the matching piano key. This process is applied on an entire video, which makes a global inference over the notes. Running this algorithm on multiple videos, in turn, enables the creation of a large dataset of pieces and their fingering information.
 
The final output we produce is demonstrated in Figure \ref{fig:finger_detection}, where we colored both the fingers and the played notes based on the pose-estimation model (\S\ref{sec:dataset:pose}) and the predicted fingers that played them (\S\ref{sec:dataset:press}). Note that the ring fingers of both hands as well as the index finger of the left hand and the middle finger of the right hand do not press any note in this particular frame, but may play a note in others. We get the information of played notes from the MIDI events.

\subsection{Data Source}
\label{sec:dataset:source}
We extract videos from \url{youtube.com}, played by different piano players on a specific channel containing both video and MIDI files. In these videos, the piano is filmed in an angle horizontal to the keyboard, from which both the keyboard and hands are displayed (as can be seen in Figure \ref{fig:finger_detection}).

\noindent\textbf{MIDI files}: 
Musical Instrument Digital Interface (MIDI) is a standard format for the interchange of musical information between electronic musical instruments. It consists of a sequence of events describing actions to carry out. 
In the setup of piano recording, it records what note was played in what time, for how long and its pressure strength (velocity). We only use videos that come along with MIDI files, as we use it as the source for the played notes and their timestamp.\footnote{A potential method for increasing the amount of data would be to use a Wav2MIDI system, which converts audio signals to MIDI, but these systems achieve poor performances on real-world scenarios \cite{DBLP:journals/corr/abs-1810-12247,DBLP:journals/corr/abs-1811-03271,cwitkowitz2019end}.}

\subsection{Keyboard and Boundaries Detection}
\label{sec:dataset:keyboard}
To allow a correct fingering assignment, we first have to find the keyboard and the bounding boxes of the keys. 
We detect the keyboard as the largest continuous bright area in the video and identify key boundaries using standard image processing techniques, taking into account the expected number of keys their predictable location and clear boundaries. For robustness and in order to handle the interfering hands that periodically hide parts of the piano, we combine information from multiple random frames by averaging the predictions from each frame.

\subsection{Hands Detection}\label{sec:dataset:hand-detection}
A straightforward approach for getting fingers locations in an image is to use a pose estimation model directly on the entire image. In practice, common methods for full-body pose estimation such as OpenPose \cite{cao2017realtime} containing hand pose estimation \cite{simon2017hand}, make assumptions about the wrist and elbow locations to automatically approximate the hands' locations. In the case of piano playing, the elbow does not appear in the videos, therefore these systems don't work.
Instead, we turn to a two-steps approach where we start by detecting the hands, crop them, and pass the hand frames to a pose estimation model.

Object Detection \cite{viola2001rapid,redmon2016you,lin2017feature,lin2017focal}, and specifically Hand Detection \cite{kolsch2004robust,sridhar2015fast,simon2017hand} are well studied subjects. 
However, we found results to be hard to replicate, due to missing code, compilation problems, private datasets and, most importantly, different lenient assumptions (e.g. no separation between left and right hands).

Therefore, we created a small dataset with random frames from different videos, corresponding to 476 hands in total, evenly split between left and right\footnote{The data was labeled by the first author using \emph{labelImg} \cite{labelimg}.}. We then fine-tuned a pre-trained object detection model (Inception v2 \cite{ioffe2015batch}, based on Faster R-CNN \cite{ren2015faster}, trained on COCO challenge \cite{lin2014microsoft}) on our new dataset.
The fine-tuned model works well and some hand detection bounding boxes are presented in Figure \ref{fig:finger_detection}.

Having an accurate hand-detection model, we perform hand detection on every frame in the video. If more than two hands are detected, we choose the two bounding boxes with the highest probability. When two hands are detected with the same label (``left-left'' or ``right-right''), we discard the model's label, and instead choose the leftmost bounding box to have the label ``left'' and the other to have the label ``right'' - which is the most common position of hands on the piano.

\subsection{Finger Pose Estimation}
\label{sec:dataset:pose}
Having the bounding box of each hand is not sufficient, as in order to assign fingers to notes we need the hand's pose.
How can we detect fingers that pressed the different keys? We turn to pose estimation, a well-studied subject in computer vision \cite{wei2016convolutional,simon2017hand,DBLP:journals/corr/abs-1903-01013} in order to tackle this part.

\begin{figure*}[t]
\centering
\begin{subfigure}{0.49\linewidth}
    \includegraphics[width=\linewidth]{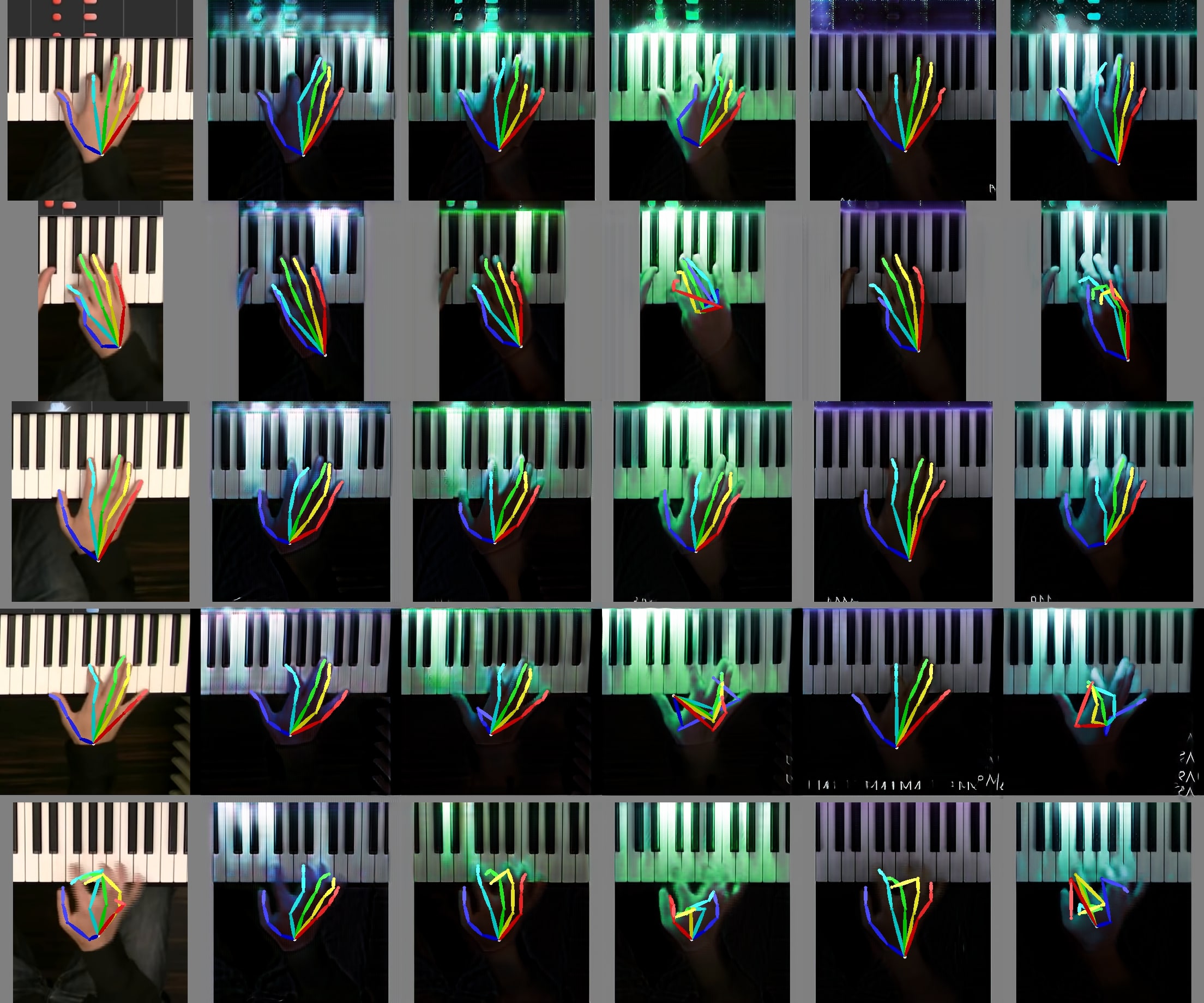}
    \caption{Pre-trained model on CycleGAN checkpoints.}
    \label{fig:cycle_gan:pretrained}
\end{subfigure}
\hfill
\begin{subfigure}{0.49\linewidth}
    \includegraphics[width=\linewidth]{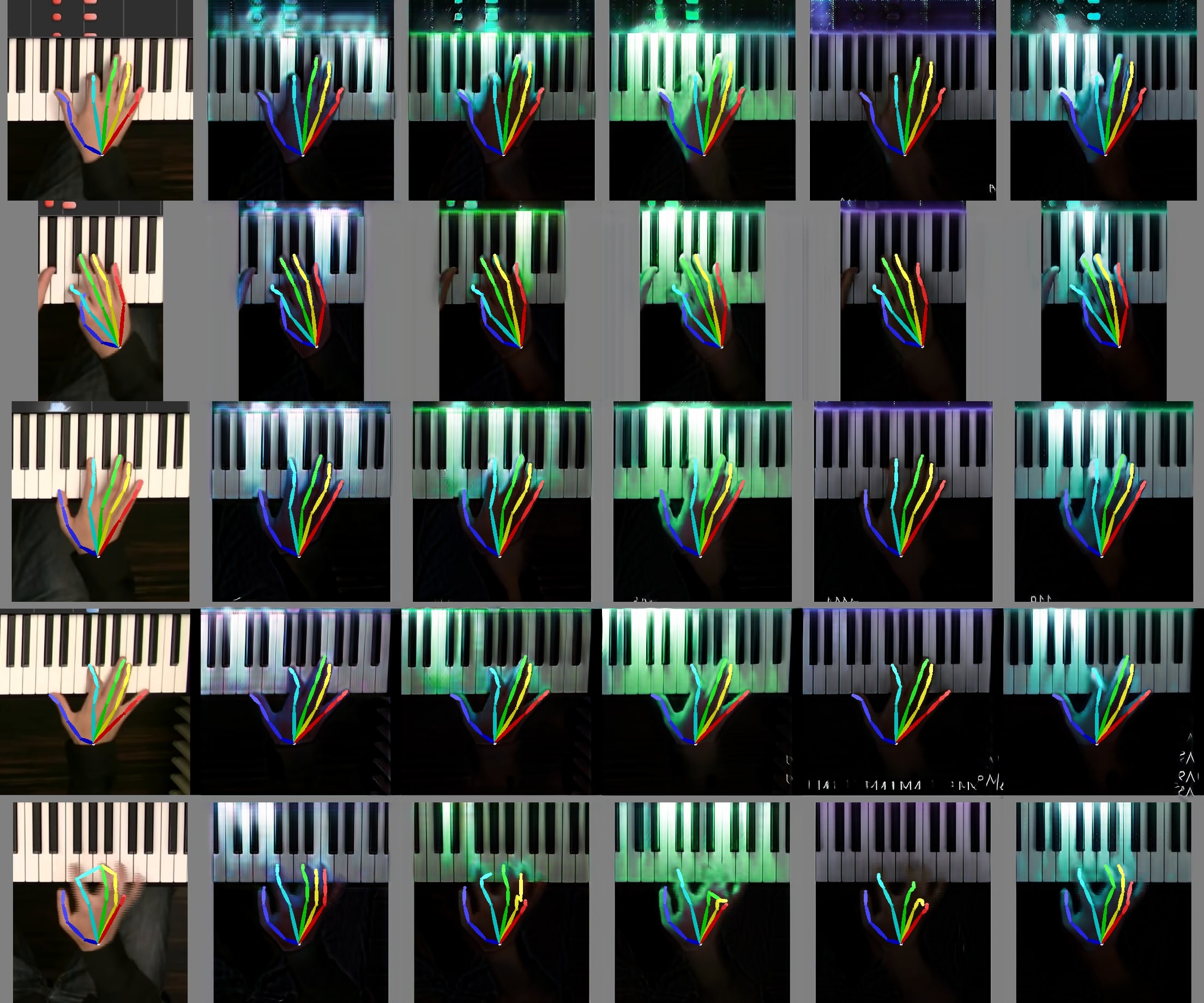}
    \caption{Fine-tuned model on CycleGAN checkpoints.}
    \label{fig:cycle_gan:finetuned}
\end{subfigure}
\begin{subfigure}{0.49\linewidth}
    \includegraphics[width=\linewidth]{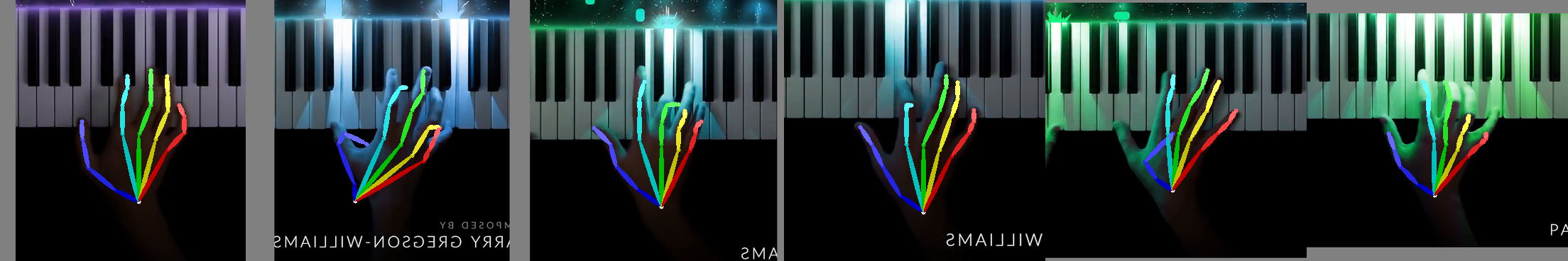}
    \caption{Pre-trained model on real-world data.}
    \label{fig:cycle_gan:pretrained-real}
\end{subfigure}
\hfill
\begin{subfigure}{0.49\linewidth}
    \includegraphics[width=\linewidth]{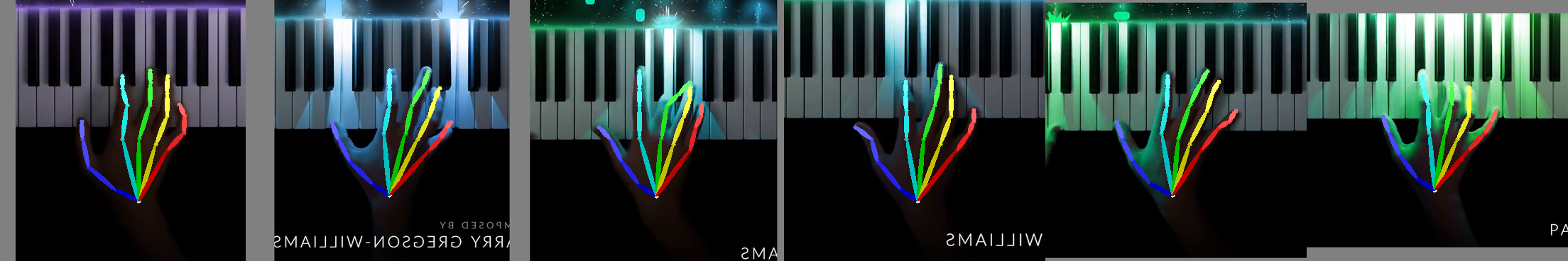}
    \caption{Fine-tuned model on real-world data.}
    \label{fig:cycle_gan:finetuned-real}
\end{subfigure}
\caption{Output of CycleGAN on 5 images, with 5 different checkpoints. The colors drawn on the hands depict their pose estimation.}
\label{fig:cycle_gan}
\end{figure*}

\noindent\textbf{Domain mismatch}:
The videos we use contain different visual effects, are usually very dark, high-contrast, and blurry due to motion blur. These are real-world conditions, which standard datasets and models rarely encounter \cite{tompson14tog,yuan2017bighand2,simon2017hand}. Therefore off-the-shelf pose estimation models often fail in our scenario. 
Some failure examples are presented in Figure \ref{fig:cycle_gan:pretrained-real} where the first pose is estimated correctly, but the rest either have wrong finger positions, shorter, or broken fingers.

\noindent\textbf{GAN-based adaptation}:
Given the observation that pose-estimation works well on well-lit images, how can we adapt it to work as well for other lighting situations? 

\input{tables/pose_estimation_domain_adaptation.tex}

We propose a new algorithm for adapting a pose-estimation model to out-of-domain inputs, described in Algorithm \ref{algo:da}.
We seek a mapping $G: S \rightarrow T$ that transforms the source domain (the well-lit videos) into the target domain (the challenging-lit scenarios). We obtain this transformation function using a CycleGAN \cite{CycleGAN2017}. Then, we use a trained pose estimation model $PE$ on the source inputs to get a silver tag $\hat{y_i}$, use the transformation $G$ to get a target-input $\hat{x_i}$, and align the prediction to the adapted input, resulting in a tuple $(\hat{x_i}, \hat{y_i})$. We use this procedure iteratively and collect silver data, which is then used to fine-tune the original pose-estimation model on. 
This algorithm results in a new model, which was fine-tuned on data from the target domain (in our case, extremely colorful images) and therefore is robust to this data.
We note that this model's execution time is equivalent to the pre-trained pose-estimation model, as we use the transformation function offline, rather then using it for every prediction. We also benefit of better generalization as we keep good performance on the source domain, and gain major performance on the target domain.

This scenario resembles to sim2real works \cite{sadeghi2018sim2real,guptacyclegan,gamrian2018transfer} where one wants to transfer a model from simulations to the real-world. 
These works learn a mapping function $G': T \rightarrow S$ that transfers instances from the target domain  $T$ (the real-world) into the source domain $S$ (the simulation), where the mapping is usually achieved by employing a CycleGan \cite{CycleGAN2017}. Then, models which are trained on the source domain are used on the transformation of the target domain $G_1(x_i)$ and manage to generalize on the target domain. 
We stress that we differ from these works conceptually as our method uses the transformation from the source to target domain, instead of the other way around. Furthermore, our method benefits from a performance boost, as at test-time, we only have to use the classification function, and not the transformation function.


We employ Algorithm \ref{algo:da} on 21,746 images we automatically collected from videos, and use the convolutional-pose-machines pose estimation \cite{wei2016convolutional}\footnote{In practice, instead of using the transformation by the CycleGAN once, we select 15 checkpoints of the model, and perform the augmentation with all of them.}. Some examples can be seen in Figure \ref{fig:cycle_gan:finetuned}, and \ref{fig:cycle_gan:finetuned-real} demonstrating the performance on different lighting scenarios.

\input{35_pressed_finger_estimation.tex}


\subsection{Video and MIDI Alignment}
\label{sec:dataset:align}
We consider the MIDI and video files to be complementary, as they were recorded simultaneously.
The MIDI file is the source for key pressing info: what key was pressed, when, and for how long. 
The video is the source for the piano, hands, and fingers locations.
These two sources are not synchronized, but as they depict the same piano performance, a perfect alignment exist (up to the video frequency resolution).

We begin by extracting the audio track from the video, and treat the first audio peak as the beginning of the piece, clipping the prior part of the video and aligning it with the first event from the MIDI file. In practice, we find the starting point as the first signal point that surpasses a threshold.

This heuristic achieves a reasonable alignment, but we observe some alignment mismatch of 80-200ms which affect the entire procedure performance.
We tackle the misalignment by using the signal from the final system confidence (Section \ref{sec:dataset:press}), where every piece gets a confidence score from our system.
We look for an alignment that maximizes the system confidence over the entire piece:
\[ 
align(MIDI, Vid) = argmax_{i}score(MIDI_{t_0}, Vid_{t_i})
\]
where $MIDI_{t_0}$ is the starting time of the MIDI file, and $Vid_{t_j}$ is the alignment time of the video. $Vid_{t_0}$ is obtained by the heuristic alignment described in the previous paragraph. We use the confidence score as a proxy of the alignment precision and search the alignment that maximizes the confidence score of the system.
More specifically, given the initial offset from the audio-MIDI alignment, we take a window of 1 second in frames (usually 25) from each side and compute the score of the final system on the entire piece. We choose the offset that results in the best confidence score as the alignment offset.


\subsection{The Resulting Dataset: \dataset}
We follow the procedure described in this section, and use it to label \pieces~piano pieces from 42 different composers with \ppnumber{\notes}~notes in total. On average, each piece contains \ppnumber{\averagenotes} notes.

%% file: tables/pose_estimation_domain_adaptation.tex
         
    

\begin{algorithm}[ht]
    \caption{Pose Estimation Domain Adaptation}
    \label{algo:da}
    \begin{algorithmic}
        \setlength\parindent{1pt}
        \State \textbf{Input:}
        unlabeled source data $\mathbb{X_S}$,
        unlabeled target data $\mathbb{X_T}$, \\
        Pose Estimation model ($PE$) for source data
        \State \textbf{Output:}
        Model for target data
        \State \textbf{Algorithm:}
        \State $L \leftarrow \{\}$
        \State Train \textbf{CycleGAN} from $\mathbb{X_S}$ to $\mathbb{X_T}$
         \For{ $x_i \in \mathbb{X_S}$ }
         	\State $\hat{y_i} = PE(x_i)$
         	\State $\hat{x_i} = \textbf{CycleGAN}(x_i)$
         	\State $L \gets (\hat{x_i}, \hat{y_i})$
         \EndFor
         
         \State Fine-Tune $PE$ on $L$
         \State Return $PE$
    
    \end{algorithmic}
\end{algorithm}

%% file: 35_pressed_finger_estimation.tex
\subsection{Pressed Finger Estimation}
\label{sec:dataset:press}

Given that we know which notes were pressed in any given frame (see \S\ref{sec:dataset:align} below), there is uncertainty as to which finger pressed them. This uncertainty either comes from imperfect pose estimation, or multiple fingers located on top of a single note.
We wish to estimate the finger press by calculating the probability of a specific finger that was used, given the hand pose and the pressed note:
$argmax_{i}~ P(x_i|n_k,pose)$

where $i \in [1, 10]$ for 10 fingers from both hands, 
$n_k \in [1, 88]$ corresponding to the played key and $pose$ correspond to the hand pose.

We model the probability of each finger being pressed as a gaussian distribution where the mean $\mu$ is the center of the key and the variance $\sigma$ is its width. Each finger gets a score for how much it ``fits'' to play the given key, calculated by the probability density function $x_{i}$, with a given finger $i$, on the key $k$ as: $g_k(x_{i}) = \mathcal{N}(x_i | \mu_k, \sigma^{2}_{k})$.
Then, to derive a distribution over fingers we normalize the scores across all fingers $P(x_{i}|n_k,pose) = \frac{g_k(x_{i})}{\sum_j g_k(x_{j,k})}$.

As most keyboard presses last more than one frame, we make use of multiple frames to overcome some of the errors from previous steps and to provide a more accurate prediction.
To achieve this, we collect the frames that were used during a key press. We treat the first frame as the main signal point, and assign each successive frame an exponentially declining weight.
Finally, we normalize the weighted sum of probabilities to achieve a probability distribution for all frames.

In our dataset, we release all probabilities for each played note, along with the maximum likelihood finger estimation. We define the ``confidence'' score of the extraction from a single piece, as the product of the highest probability for a finger for each note.

%% file: 41_results.tex
\section{Evaluation}
\label{results}

In this section, we provide two evaluations that show that (1) our system for detecting fingering from videos is accurate and (2) the dataset we achieved by repeating this process over dozens of videos is of good quality.



\subsection{Finger Press Estimation Evaluation}\label{sec:results:press-estimation}

We manually annotated five random pieces from our dataset by marking the pressing finger for each played note in the video. Then, by using our system (\S\ref{sec:dataset:press}), we estimate what finger was used for each key. We use the confidence score produced by the model as a threshold to use or discard the key prediction of the model, and report precision, recall and F1 scores of multiple thresholds. As discussed in Section \ref{sec:dataset:pose}, current pose-estimation models perform poorly, and we propose a new method for domain adaptation. In Table \ref{table:final_system_results} we report the entire system performance on the original pose-estimation \cite{wei2016convolutional} model (Pretrained) using multiple frames, as well as the results using fine-tuned, both by using just the first frame (FT Single Frame) and multiple frames (FT Multiple Frames), which performs best. 


When considering a high confidence score (\textgreater{}90\%) both the pre-trained and fine-tuned models correctly mark all of the considered notes (which consist of between 34-36\% of the data). However, when considering decreasing confidences\footnote{For brevity we consider here only a threshold of 0.5, but we observe the same phenomenon with more thresholds.}, our method achieves higher precision and higher recall. With no confidence threshold (i.e using all fingering predictions), the pre-trained model achieves 93\% F1, while the fine-tuned one achieves 97\% F1, a 57\% error reduction.

\input{tables/pose_estimation_results.tex}

\subsection{Automatic Piano Fingering Prediction}\label{sec:results:fingering-prediction}

In Section \ref{sec:results:press-estimation} we showed that our method for extracting \fingering{} from videos is accurate. By running this process on dozens of videos we created the largest dataset for \fingering{} to date. Is this dataset useful?

We begin by training a standard sequence tagging model on \dataset~and perform evaluations on the subset we manually annotated and on \pig{} \cite{nakamura2019statistical}. Then, by fine-tuning the pre-trained model on \pig{}, we achieve improvements in performance over the previous state-of-the-art which was trained solely on \pig{}, suggesting that our dataset is indeed accurate and useful.


We model the \fingering{} as a sequence labeling task, where given a sequence of notes $n_1,n_2,...,n_n$ we need to predict a sequence of fingering: $y_1,y_2,...,y_n$, where $y_i \in \{1,2,3,4,5\}$ corresponding to 5 fingers of one hand. We employ a standard sequence tagging technique, by embedding each note and using a BiLSTM on top of it. On every contextualized note we then use a Multi-Layer-Perceptron (MLP) to predict the label.
The model is trained to minimize cross-entropy loss. This is the same model used in \cite{nakamura2019statistical}, referred to as DNN (LSTM).

For evaluation, we use the match rate between the prediction and the ground truth. For cases where there is a single ground truth, this is equivalent to accuracy measurement. When more than one labeling is available, we simply average the accuracies with each labeling.\footnote{This matches the \textit{general match rate} evaluation metric in \cite{nakamura2019statistical}.}
The results are summarized in Table \ref{tbl:apfd_pig_results}.

\input{tables/fingering_results.tex}

We begin by experimenting on our dataset.
\dataset{} is composed of \pieces{} pieces, which we split into 75/10 for training and development sets respectively, and use the 5 manually annotated pieces as a test set. We note that the development set is silver data (automatically annotated) and is likely to contain mistakes. We run the model and achieve 73.2\% accuracy.

Next, to test how useful our data is to a real-world gold dataset we wish to inspect its usefulness with a transfer-learning approach. By first pre-training a simple model on our data and then fine-tune the model on \pig{}. If our data is indeed of quality we'd expect to see performance improvements on \pig{}, especially as this data is relatively small.

We begin by training the LSTM model solely on \pig{},\footnote{\cite{nakamura2019statistical} didn't use a development set, therefore in this work, we leave 1 piece from the training set and make it a development set.} which results in 64.1\% accuracy. This result is higher than the neural model of the same architecture that was reported by \cite{nakamura2019statistical} (61.3\%) but 0.4\% lower than their best performing model (an HMM based). We continue by using the model trained on \dataset{} and then fine-tune it on \pig{}, which leads to 66.8\% accuracy, an improvement of 2.7 points over the previous SOTA. We attribute this gain in performance to our dataset, which both increases the number of training examples and allows to train bigger neural models which excel with more training examples.
We also experiment in the opposite direction and fine-tune the model trained on PIG with our data, which result in 73.6\% accuracy, which is better than training on our data alone, achieving 73.2\% accuracy.


%% file: tables/pose_estimation_results.tex


\begin{table}[t]
\resizebox{\columnwidth}{!}{%
\begin{tabular}{l|ccc|ccc|ccc}
\toprule
Confidence & \multicolumn{3}{c|}{c \textgreater{} 0.9} & \multicolumn{3}{c|}{c \textgreater{} 0.5} & \multicolumn{3}{c}{c \textgreater{} 0}\\
Metrics & p & r & f1 & p & r & f1 & p & r & f1\\
\midrule
Pretrained & \textbf{1.0} & 0.34 & 0.51 & 0.95 & 0.78 & 0.86 & 0.88 & \textbf{1.0} & 0.93\\
FT Single Frame & 0.99 & \textbf{0.36} & \textbf{0.53} & 0.97 & \textbf{0.8} & \textbf{0.88} & 0.9 & \textbf{1.0} & 0.95\\
FT Multiple Frames & \textbf{1.0} & 0.35 & 0.52 & \textbf{0.98} & \textbf{0.8} & \textbf{0.88} & \textbf{0.94} & \textbf{1.0} & \textbf{0.97} \\
\bottomrule
\end{tabular}
}

\caption{Precision, Recall, F1 scores for 6,894 notes we manually tagged across 5 different pieces, at different confidence points. The first row depict the performance of the system with the vanilla pose estimation. The second and the third rows present the results on the fine-tuned (FT) pose estimation model (\S\ref{sec:dataset:pose}) on a single and multiple frames accordingly.}
\label{table:final_system_results}
\end{table}

%% file: tables/fingering_results.tex


\begin{table}[t]
\centering
\begin{tabular}{l|ll}
Training & \multicolumn{2}{c}{Evaluation Data} \\
  & \dataset  & PIG           \\ \hline \hline
Human Agreement   & ---  & 71.4          \\ \hline
Previous Neural        & ---  & 61.3          \\
Previous SOTA         & ---  & 64.5          \\ \hline
PIG               & 49.9 & 64.1          \\
~ + FT \dataset & \textbf{73.6} & 54.4          \\ \hline
\dataset  & 73.2 & 55.2          \\
~ + FT PIG    & 62.9 & \textbf{66.8}
\end{tabular}
\caption{Results of training and fine tuning on \dataset{} and \pig{}.}
\label{tbl:apfd_pig_results}
\end{table}

%% file: 51_discussion.tex
\section{Discussion and Future Work}
In this work, we present an automatic method for detecting \fingering{} from MIDI and video files of a piano performance. We employ this method on a large set of videos, and create the first large scale \fingering{} dataset, containing \pieces{} unique pieces, with \ppnumber{\notes~} notes in total. We show that this dataset---although noisy--is valuable, by pre-training a model on it and fine-tuning it on a gold dataset, where we achieve state-of-the-art results. In future work, we intend to improve the data collection by ameliorating the pose-estimation model to better handle the high speed movements and the proximity of the hands, which often cause errors in estimating their pose. Furthermore, we intend to design improved neural models that can take previous fingering predictions into account, in order to have a better global fingering transition.

%% file: 999_appendix.tex
\section{Dataset Samples}
For every video our system extracts \fingering~from it also outputs the video overlayed by the estimation of the piano keys, an indication of which notes are played, and what fingers are being used to play them (the key's color), up to two bounding boxes, for both hands (where light blue means left hand, and light green means right hand), and the pose estimation for each hand.

We include the output videos for all of the pieces we manually annotated to visually demonstrate our system and its accuracy on different playing cases. The videos were uploaded to a new, anonymous YouTube channel, and each contain a link to the original video in the description.

\begin{itemize}
    \item Yiruma - River Flows in You: \\ \url{https://youtu.be/Gfs1UWQhr5Q}
    \item Alan Walker - Faded: \\ \url{https://youtu.be/LU2ibOW6z7U}
    \item Beethoven - Moonlight Sonata 1st Movement: \\ \url{https://youtu.be/wp8j239fs9o}
    \item Mozart - Rondo Alla Turca: \\ \url{https://youtu.be/KqTaPfoIuuE}
    \item Chopin - Nocturne in E Flat Major (Op. 9 No. 2): \\ \url{https://youtu.be/xXHUUzTa5vU}
\end{itemize}

\begin{figure}[h]
    \centering
    \includegraphics[width=0.8\columnwidth]{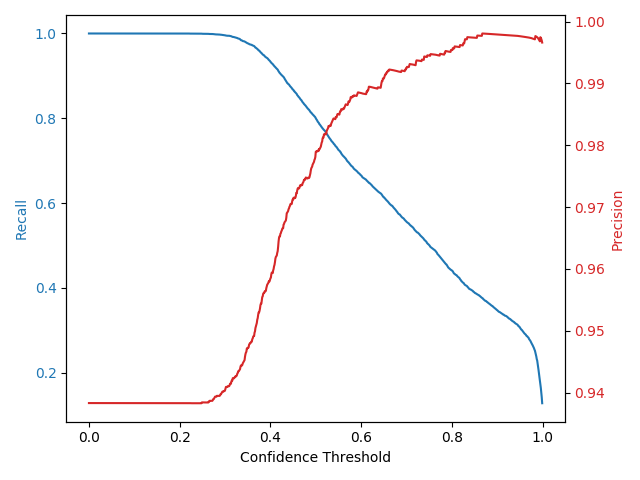}
    \caption{Precision and recall based on confidence for all 6,894 manually tagged notes}
    \label{fig:smooth_confidence}
\end{figure}